\pdfoutput=1
\documentclass[11pt]{article}

\usepackage[]{acl}

\usepackage{times}
\usepackage{latexsym}
\usepackage{listings}
\usepackage{hyperref}
\usepackage[T1]{fontenc}
\usepackage[utf8]{inputenc}
\usepackage{amsfonts}
\usepackage{microtype}

\usepackage{inconsolata}

\usepackage{graphicx}

\usepackage{booktabs}
\usepackage{multirow}
\usepackage{xcolor}
\definecolor{darkgreen}{rgb}{0.0, 0.5, 0.0}
\usepackage{pgfplots}
\usetikzlibrary{positioning}
\usepgfplotslibrary{groupplots}
\pgfplotsset{compat=1.18}
\usepackage{subcaption}
\usepackage{amsmath}
\usepackage{cleveref}
\usepackage{amssymb}

\usepackage[most]{tcolorbox}
\tcbuselibrary{breakable}
\usepackage{listings}
\usepackage{hyperref}
\usepackage{geometry}
\usepackage{algorithm}
\usepackage{algorithmic}

\definecolor{model1}{HTML}{4287f5}  % Blue
\definecolor{model2}{HTML}{42c5f5}  % Light Blue
\definecolor{model3}{HTML}{f542a7}  % Pink
\definecolor{model4}{HTML}{8c42f5}  % Purple
\definecolor{model5}{HTML}{42f5c5}  % Teal
\definecolor{model6}{HTML}{f5a742}  % Orange
\title{Error Typing for Smarter Rewards: Improving Process Reward Models with Error-Aware Hierarchical Supervision}

\author{
    \textbf{Tej Deep Pala\textsuperscript{1}},
    \textbf{Panshul Sharma\textsuperscript{1}}\\
    \textbf{Amir Zadeh\textsuperscript{2}},
    \textbf{Chuan Li\textsuperscript{2}},
    \textbf{Soujanya Poria\textsuperscript{1}}\\\\
    \textsuperscript{1}Singapore University of Technology and Design\\
    \textsuperscript{2}Lambda Labs\\
  }

\newcommand{\model}[1]{\texttt{PathFinder-PRM#1}}

\newtcolorbox{promptbox}[1][]{
  enhanced,
  breakable,                    
  colback=gray!5,
  colframe=gray!50!black,
  width=\linewidth,
  boxsep=1mm,                   
  left=2mm, right=2mm,          
  top=1mm, bottom=1mm,          
  before skip=6pt,              
  after skip=6pt,               
  title=#1,
  attach boxed title to top left={yshift=-2mm,xshift=2mm},
  boxed title style={
    size=small,
    colback=gray!50!black,
    colframe=gray!50!black,
    boxsep=1mm,
  },
}

\begin{document}
\maketitle
\begin{abstract}
Large Language Models (LLMs) are prone to hallucination, especially during multi‑hop and reasoning-intensive tasks such as mathematical problem solving. While Outcome Reward Models verify only final answers, Process Reward Models (PRMs) score each intermediate step to steer generation toward coherent solutions. We introduce \model{}, a novel hierarchical, error‑aware discriminative PRM that first classifies math and consistency errors at each step, then combines these fine‑grained signals to estimate step correctness. To train \model{}, we construct a 400K‑sample dataset by enriching the human‑annotated PRM800K corpus and RLHFlow Mistral traces with three‑dimensional step‑level labels. On PRMBench, \model{} achieves a new state‑of‑the‑art PRMScore of 67.7, outperforming the prior best (65.5) while using 3× less data. When applied to reward guided greedy search, our model yields prm@8 48.3, a +1.5 point gain over the strongest baseline. These results demonstrate that decoupled error detection and reward estimation not only boost fine‑grained error detection but also substantially improve end‑to‑end, reward‑guided mathematical reasoning with greater data efficiency~\footnote{Code: https://github.com/declare-lab/PathFinder-PRM}.

\end{abstract}

\section{Introduction}

Large language models (LLMs) have achieved remarkable success on many natural language tasks, including open‑ended generation and complex reasoning \cite{Brown2020Fewshot, Wei2022COT}. However, they remain prone to \emph{hallucinations} and subtle logical errors when generating multi‑step solutions, particularly in domains such as mathematical problem solving \cite{wang2023selfconsistencyimproveschainthought, zheng2024processbenchidentifyingprocesserrors}. Traditional outcome‐only verifiers (Outcome Reward Models) can check a final answer but fail to catch intermediate missteps that lead reasoning astray \cite{Wang2024}.

To address this gap, \emph{Process Reward Models} (PRMs) have been proposed, which assign individual rewards to each reasoning step. \cite{uesato2022solvingmathwordproblems,liu2025can}. As such, PRMs can filter out erroneous chains of thought and guide generation toward more reliable reasoning trajectories \cite{lightman2023lets, zhang2025lessons}. 

Recent interest in explicitly reasoning-centric LLMs such as DeepSeek-R1 and OpenAI’s GPT-o series models underscores the field’s growing emphasis on human-like thinking and the ability to flexibly scale test-time compute \cite{guo2025deepseek,openai2025o3o4}. These models demonstrate extended deliberation and use structured reasoning traces to solve complex problems. In such settings, effective \emph{process supervision} is crucial: rather than merely verifying a final answer, it must guide and correct the reasoning process at every step, ensuring logical coherence and factual accuracy throughout. PRMs are thus an essential component in aligning reasoning LLMs with reliable multi-step reasoning.

Despite recent advances, current PRMs still struggle with fine‑grained error types. For example, the PRMBench benchmark reveals that many state‑of‑the‑art PRMs fall short of detecting subtler faults such as non‑redundancy violations, domain inconsistencies, or deceptive logical steps \cite{song2025prmbench}. Moreover, existing methods typically combine \emph{error detection} (is this step wrong?) with \emph{path optimality} (how helpful is this step in reaching the solution?) in a single prediction, leaving each signal underutilized \cite{zhang2025lessons,xia2025evaluating}.

\begin{figure*}[ht]
    \centering
    \includegraphics[width=0.7\linewidth]{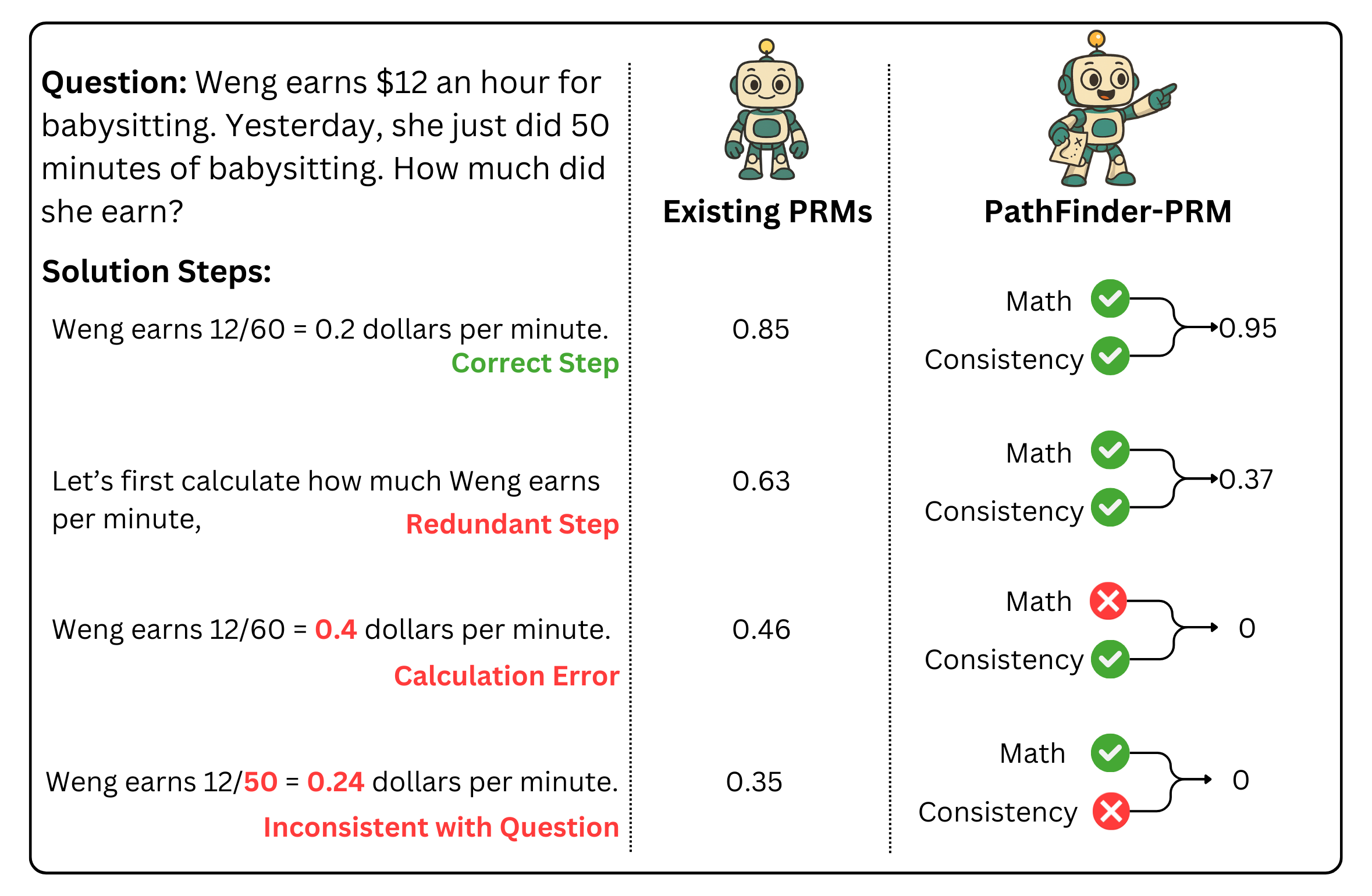}
    \caption{Comparing the Methodology of Existing PRMs against \model{}.}
    \label{fig:Pipeline}
\end{figure*}

In this work, we argue that error detection and value estimation are complementary but distinct objectives. By decoupling them into two sequential subtasks, first explicitly identifying specific error categories and using those error signals to compute a step‑level reward, we can obtain richer supervision and stronger guidance for downstream generation (depicted in \Cref{fig:Pipeline}). To this end, we introduce \model{}, an error-aware hierarchical PRM that (1) classifies each step to detect the presence of math or consistency errors, and (2) combines these fine‑grained error labels to produce a final reward score.

We construct a new training corpus by augmenting the human-annotated \textit{PRM800K} \cite{lightman2023lets} and automated RLHFlow Mistral data \cite{xiong2024rlhflowmath} with our three‑dimensional labels, yielding around 400K richly annotated reasoning trajectories. 

Our experiments on ProcessBench \cite{zheng2024processbenchidentifyingprocesserrors}, PRMBench \cite{song2025prmbench}, and a suite of end‑to‑end math benchmarks show that \model{} not only establishes the new state of the art among PRMs trained on PRM800K‑only data but also continues to scale gracefully when incorporating additional automated annotations.

\paragraph{Contributions}
\begin{itemize}
  \item We propose \model{}, the first hierarchical PRM that explicitly types errors into \emph{math} and \emph{consistency} categories before estimating step optimality.
  \item We curate a multi‑source dataset of 400K mathematical reasoning trajectories with three‑dimensional step‑level labels, combining PRM800K and RLHFlow data under our unified schema.
  \item We demonstrate that error‑aware hierarchical supervision yields substantial gains on ProcessBench and PRMBench, and leads to more accurate and robust end‑to‑end math problem solving under reward‑guided search.
  
\end{itemize}

\section{Related Works}
\textbf{Process Reward Models: } Process reward models (PRMs) evaluate the quality of intermediate steps in reasoning processes. PRMs are a core component of test-time scaling, enabling small policy models to outperform much larger models on reasoning tasks by reinforcing productive reasoning pathways \cite{liu2025can}. Most recent work on PRMs has varied along two primary axes: (1) the choice of the base model architecture and (2) the methods used to synthesize step-level supervision labels \cite{Wang2024, zhang2025lessons, khalifa2025}. Discriminative PRMs are designed to classify the correctness of individual process steps within reasoning trajectories \cite{uesato2022solvingmathwordproblems}. Corresponding training data synthesis approaches include gold human step-level annotations \cite{lightman2023lets}, Monte Carlo estimation \cite{Wang2024, luo2024improvemathematicalreasoninglanguage}, and consensus filtering methods \cite{zhang2025lessons}. Generative PRMs scale verifier compute, utilizing the language modeling head of the PRMs to generate a chain of thought (CoT) before producing step-level correctness classifications. Associated training data is typically obtained by filtering LLM-as-a-judge reasoning traces against human step-level labels \cite{khalifa2025, she2025r} or relative progress estimation \cite{zhao2025genprm}. 

\textbf{Limitations of Current PRMs: } Current PRMs face significant challenges in detecting nuanced error types. Although frontier models excel at identifying obvious mistakes \cite{zheng2024processbenchidentifyingprocesserrors}, their performance deteriorates markedly when confronted with more subtle error types. The recently introduced \texttt{PRMBench} benchmark demonstrates this limitation, revealing substantial performance drops across fine-grained error categories such as redundancy, circular logic, step inconsistency and domain inconsistency \cite{song2025prmbench}.

\section{Methodology}

\subsection{Task: Math Process Reward Modeling}

The primary task of a Process Reward Model in mathematical problem solving is to evaluate the correctness of each intermediate reasoning step generated by an LLM. Unlike ORMs, which assess only the final answer, PRMs evaluate at the step level, enabling fine-grained supervision and improved interpretability.

Formally, given a multi-step solution $\mathcal{S} = \{s_1, s_2, \ldots, s_T\}$ produced by a language model for a math problem $Q$, the PRM assigns a scalar reward $r_t \in \mathbb{R}$ to each step $s_t$, reflecting its mathematical and logical correctness. These rewards serve two main purposes:

\begin{enumerate}
    \item \textbf{Error Detection:} By evaluating each step individually, PRMs can identify inaccurate or hallucinated reasoning in the solution process, even when the final answer may coincidentally be correct.
    
    \item \textbf{Guidance Toward Correct Solutions:} The step-level feedback provided by PRMs can be used to steer generation policies, in reinforcement learning or reward-guided generation, towards valid and logically coherent trajectories, thereby improving the overall quality of generated solutions.
\end{enumerate}

\noindent Process Reward Models enhance the robustness of mathematical reasoning systems by aligning training and inference with process-based correctness, rather than relying only on outcome validation.

\subsection{\model{}}

Existing Process Reward Models tackle both error detection and optimal path guidance jointly. Given a mathematical problem $Q$ and a sequence of solution steps generated by a policy model, $\mathcal{S} = \{s_1, s_2, \ldots, s_T\}$, these models assign each step a reward score conditioned on the question and all previous steps:

\begin{equation*}
R_t = PRM\bigl(Q, \mathcal{S}_{<t}, s_t\bigr).
\end{equation*}

where $\mathcal{S}_{<t} = \{s_1, s_2, \ldots, s_{t-1}\}$
The resulting score implicitly reflects both the presence of errors in $s_t$ and its contribution towards a correct solution.

In contrast, our approach takes a hierarchical perspective by decomposing the reward assignment into two sequential subtasks: (a) detecting errors in each reasoning step, and (b) using the error information to inform step optimality. Specifically, we first categorize errors into two types:

\begin{enumerate}

\item \textbf{Math Errors:} mistakes in arithmetic or algebraic manipulation, incorrect formula application, or invalid implications.

\item \textbf{Consistency Errors:} logical inconsistencies with the question, prior steps, or established constraints.

\end{enumerate}

As seen in \Cref{fig:Pipeline}, \model{} performs two forward passes per step. In the first pass, it predicts the correctness probabilities for the two error types, $M_t$ and $C_t$. In the second pass, it estimates the step's reward $R_t$, explicitly conditioned on the detected error labels:

\resizebox{0.9\linewidth}{!}{%
  \begin{minipage}{\linewidth}
    \begin{align*}
      M_t, C_t =\ &\hspace{0.5em} \text{\model{}}\bigl(Q, \mathcal{S}_{<t}, s_t\bigr), \\
      R_t       = &\hspace{0.5em} \text{\model{}}\bigl(Q, \mathcal{S}_{<t}, s_t, M_t, C_t\bigr).
    \end{align*}
  \end{minipage}
}

By structuring reward modeling hierarchically, \model{} leverages fine-grained error signals to improve reward estimation, while maintaining a clear separation between error detection and correct path guidance. We provide further details regarding the inference design in \Cref{sec: inference design}

To investigate this, we propose the following hypothesis:\\
\fbox{\parbox{\linewidth}{\textbf{Hypothesis:} A hierarchical supervision strategy—first detecting error types, then using them to compute rewards—is more effective than existing methods that compute rewards directly without identifying the presence of errors explicitly.}}

\subsection{Creating the \model{} Dataset}

The \texttt{PathFinder-PRM} dataset includes step-level fine-grained labels across three categories: (1) \emph{mathematical reasoning accuracy}, (2) \emph{consistency} with prior steps and mathematical domain, and (3) \emph{step correctness}. Here, the third category \emph{step correctness} identifies whether a step is both error-free and optimally contributes to solving the problem. For each process step, we assign a three-dimensional categorical score vector $\mathbf{c}_t = (c_t^{\text{math}}, c_t^{\text{consistency}}, c_t^{\text{correctness}})$, where each component $c_t^{(i)} \in \{0, 1\}$ represents a binary label for the respective category. We construct this dataset by leveraging two existing datasets: the PRM800k \cite{lightman2023lets} and Mistral-PRM-Data by RLHFlow \cite{xiong2024rlhflowmath}. These datasets contain step-level correctness annotations generated through human evaluation and Monte Carlo estimation, respectively.

\textbf{PRM800K Integration:} The original PRM800k dataset contains over 800,000 gold step-level correctness labels $l_t \in \{-1, 0, 1\}$ for each reasoning step $s_t$. We transform each correctness label into our categorical score vector as follows: 
\begin{itemize}
    \item Steps with $l_t = 1$ (correct and optimal) are mapped to $\mathbf{c}_t = (1, 1, 1)$.
    \item Steps with $l_t = 0$ (correct but suboptimal) are mapped to $\mathbf{c}_t = (1, 1, 0)$.
\end{itemize}
This mapping reflects our interpretation that human labels $l_t \in \{0, 1\}$ indicate error-free reasoning, with $l_t = 0$ specifically denoting non-optimal process steps. For erroneous steps ($l_t = -1$), the original correctness labels provide insufficient information to determine scores across our three evaluation categories. Therefore, we employ DeepSeek-R1-Distill-Qwen-32 B to generate binary labels for each category for these steps \cite{deepseekai2025deepseekr1incentivizingreasoningcapability}. To maintain dataset quality, we subsequently filter out samples with categorical score vectors that are inconsistent with $-1$ human annotated labels (i.e., $\mathbf{c}_t = (1, 1, 1)$).

\textbf{Mistral-PRM-Data Integration:} Since this dataset lacks gold standard step-level correctness labels, we utilize DeepSeek-R1-Distill-Qwen-32B to assign binary categorical labels to a small, randomly selected subset of process steps. To ensure data quality, we implement a consistency filtering mechanism. This removes score assignments that are logically incompatible with the existing Monte Carlo (MC) estimation labels. Specifically:

\begin{itemize}
    \item For steps with MC estimation '+' labels (indicating positive assessment), we retain only samples with assignments of  $\mathbf{c}_t = (1, 1, 1)$.
    \item For steps with MC estimation '-' labels (indicating negative assessment), we retain samples with categorizations where at least one component equals 0, i.e., $\mathbf{c}_t \neq (1, 1, 1)$.
\end{itemize}

In totality, the \model{} dataset contains about 400K reasoning trajectory samples with step-level categorical score vectors $\mathbf{c}_t$. Of these 400K trajectories, approximately 345K are sourced from PRM800k and the other 55K reasoning paths are sourced from Mistral-PRM-Data. We train two variants of the model, \model{-7B} and \model{-7B-PRM800k} trained on the full dataset and just the PRM800K subset respectively.

\subsection{Training Recipe for \model{}}

Previous studies demonstrate that a model’s mathematical reasoning ability correlates with its performance as a process reward model~\cite{xia2025evaluating}. Consequently, we initialize \model{} from \texttt{Qwen2.5-Math-7B-Instruct}, which achieves state-of-the-art results on multiple math benchmarks~\cite{yang2024qwen25mathtechnicalreportmathematical}. Unlike recent PRMs that swap the language modeling head for a scalar value head~\cite{zhang2025lessons, xia2025evaluating, tan2025aurora}, we preserve the original LM architecture and extend the tokenizer with two special tokens, \texttt{<+>} and \texttt{<->}, to represent positive and negative step labels.

\paragraph{Training Objective}
Each training example is structured in two parts, mirroring the inference passes:

\begin{enumerate}
\item \emph{Error Detection Target:} 
\begin{quote}
Prompt + Math: \texttt{<+>}/\texttt{<->}, Consistency: \texttt{<+>}/\texttt{<->}
\end{quote}

\item \emph{Reward Estimation Target:} Append the predicted error labels and the token
\begin{quote}
Prompt + Math: [Math label], Consistency: [Consistency Label] + Correctness: \texttt{<+>}/\texttt{<->}
\end{quote}
\end{enumerate}

For each sample, we compute the cross-entropy loss only on these label tokens.

\section{Experimental Setup}

\begin{table*}[ht!]
\centering
\resizebox{\textwidth}{!}{%
    \begin{tabular}{@{}l
      *{3}{c}
      *{5}{c}
      *{4}{c}
      c@{}}
    \toprule
    \multirow{2}{*}{\textbf{Model}}
      & \multicolumn{3}{c}{\textbf{Simplicity}}
      & \multicolumn{5}{c}{\textbf{Soundness}}
      & \multicolumn{4}{c}{\textbf{Sensitivity}}
      & \multirow{2}{*}{\textbf{Overall}} \\
    \cmidrule(lr){2-4} \cmidrule(lr){5-9} \cmidrule(lr){10-13}
      & \textbf{NR}.  & \textbf{NCL}. & \textbf{Avg}.
      & \textbf{ES}.   & \textbf{SC}.  & \textbf{DC}.   & \textbf{CI}.   & \textbf{Avg}.
      & \textbf{PS}.   & \textbf{DR}.  & \textbf{MS}.   & \textbf{Avg}. & \\
    \midrule
    \multicolumn{14}{l}{\textbf{{LLM‑as‑judge, Open-source Language Models}}} \\
    \midrule
    Qwen-2.5-Math-72B*      & 55.3 & 54.9 & 55.1 & 55.5 & 71.6 & 58.1 & 59.1 & 61.1 & 47.4 & 53.8 & \textbf{100.0} & 67.1 & 57.4 \\
    QwQ-Preview-32B*       & \textbf{57.2} & \textbf{55.6} & \textbf{56.4} & \textbf{67.4} & \textbf{72.3} & \textbf{66.2} & \textbf{66.9} & \textbf{68.2} & \textbf{57.8} & \textbf{62.7} & \textbf{100.0} & \textbf{73.5} & \textbf{63.6} \\
    % R1-Distill-Llama3.1-70B*     & 49.5 & 48.1 & 48.8 & 61.4 & 65.5 & 65.8 & 61.1 & 63.4 & 48.8 & 54.1 & \textbf{100.0} & 67.6 & 57.5 \\
    % DeepSeek-R1 (671B)*        & 63.0 & 62.7 & 62.9 & 68.2 & 68.5 & 73.5 & 75.4 & 71.4 & 63.3 & 68.0 & 100.0 & 77.1 & 67.8 \\
    \midrule
    \multicolumn{14}{l}{\textbf{{LLM‑as‑judge, Proprietary Language Models}}} \\
    \midrule
    GPT‑4o*       & 57.0 & 62.4 & 59.7 & 72.0 & 69.7 & 70.7 & 71.1 & 70.9 & \textbf{62.5} & \textbf{65.7} & 99.2 & \textbf{75.8} & 66.8 \\
    Gemini-2.0-flash-exp*    & 67.2 & 58.1 & 62.7 & 70.4 & 65.7 & 66.0 & 67.3 & 67.3 & 61.8 & 66.2 & 98.2 & 75.4 & 66.0 \\
    Gemini-2.0-thinking-exp-1219*    & \textbf{68.5} & \textbf{63.8} & \textbf{66.2} & \textbf{72.9} & \textbf{71.3} & \textbf{71.0} & \textbf{71.8} & \textbf{71.8} & 60.3 & \textbf{65.7} & \textbf{99.8} & 75.3 & \textbf{68.8} \\
    \midrule
    \multicolumn{14}{l}{\textbf{{Discriminative Process Reward Models}}} \\
    \midrule
    Math‑Shepherd‑7B*     & 44.0 & 50.3 & 47.1 & 49.4 & 44.5 & 41.3 & 47.7 & 45.7 & 47.2 & 48.6 & 86.1  & 60.7 & 47.0 \\
    Math‑PSA‑7B$^\dag$           & 47.6 & 55.1 & 51.3 & 56.5 & 49.4 & 47.1 & 54.2 & 51.8 & 51.7 & 54.1 & 88.9  & 64.9 & 52.3 \\
    RLHFlow‑Mistral‑8B*   & 46.1 & 47.3 & 46.7 & 56.6 & 55.1 & 54.4 & 63.8 & 57.5 & 51.5 & 56.2 & 97.9  & 68.5 & 54.4 \\
    RLHFlow‑DeepSeek‑8B*  & 46.4 & 48.9 & 47.6 & 55.7 & 55.0 & 53.2 & 66.2 & 57.5 & 49.0 & 55.4 & \textbf{99.8} & 68.1 & 54.2 \\
    Lemma‑PRM800k‑7B*     & 49.3 & 53.4 & 51.4 & 56.4 & 47.1 & 46.7 & 53.3 & 50.9 & 51.0 & 53.5 & 93.6  & 66.0 & 52.0 \\
    Skywork‑PRM‑7B*       & 35.7 & 41.2 & 38.4 & 36.7 & 29.1 & 30.6 & 34.4 & 32.7 & 36.8 & 37.4 & 88.8  & 54.3 & 36.2 \\
    ReasonEval‑7B*        & \textbf{61.0} & 50.1 & 55.5 & 62.1 & 65.9 & 61.5 & 66.0 & 63.9 & 55.6 & 58.0 & 99.5  & 71.0 & 60.0 \\
    Qwen2.5‑Math‑7B‑PRM800K$^\dag$ & 48.6 & 47.8 & 48.2 & 62.1 & 59.4 & 58.7 & 68.5 & 62.2 & 52.9 & 64.0 & \textbf{99.8} & 72.2 & 58.3 \\
    Qwen2.5‑Math‑PRM‑7B*    & 49.0 & 55.1 & 52.1 & 71.8 & 67.3 & 66.3 & \textbf{78.5} & 71.0 & 57.6 & 69.1 & 99.7 & 75.5 & 65.5  \\[2ex]
    $\circledast$ \model{}-7B-PRM800K & 51.5 & 61.3 & 56.4 & 69.7 & 67.6 & 65.9 & 71.9 & 68.8 & 58.7 & 66.6 & 99.4 & 74.9 & 65.0 \\
    \hspace{0.7em} w/o Separate Subtask Prediction& 58.9 & \textbf{66.7} & \textbf{62.8} & 68.6 & 62.4 & 62.4 & 66.7 & 65.0 & 60.2 & 64.9 & 97.8 & 74.3 & 64.4 \\
    \midrule
    $\circledast$ \model{}-7B & 52.1 & 65.8 & 58.9 & 73.1 & 68.7 & 66.3 & 75.0 & 70.8 & \textbf{61.7} & \textbf{69.8} & 99.2 & \textbf{76.9} & \textbf{67.7} \\
    \hspace{0.7em} w/o Separate Error Categories& 51.7 & 62.0 & 56.9 & \textbf{73.2} & \textbf{70.0} & \textbf{66.9} & 75.8 & \textbf{71.5} & 60.3 & 69.2 & 99.6 & 76.4 & 67.3 \\
    \hspace{0.7em} w/o Separate Subtask Prediction & 57.9 & 66.4 & 62.1 & 69.1 & 62.6 & 62.2 & 68.7 & 65.7 & 61.0 & 65.4 & 98.2 & 74.9 & 64.9 \\
    \bottomrule
    \end{tabular}
}
\caption{Performance on \textbf{PRMBench}. Results marked with * and \dag come from \citeauthor{song2025prmbench} and \citeauthor{she2025r} respectively. Bold text denotes the best results within each category. $\circledast$ represents the models we trained.} 
\label{tab:prmbench_main}
\end{table*}

\subsection{Evaluation Benchmarks}
For math steps error detection, we use ProcessBench and PRMBench. \textbf{ProcessBench} is a benchmark designed to evaluate language models' ability to identify errors in mathematical reasoning processes. It comprises 3,400 test cases, primarily sourced from different math reasoning benchmarks. Each case includes a step-by-step solution annotated by human experts to indicate the earliest step containing an error or to confirm the correctness of all steps. Models are tasked with pinpointing the first erroneous step in a solution or affirming the solution's correctness. \textbf{PRMBench} is a fine-grained benchmark aimed at evaluating Process-Level Reward Models (PRMs) on their capability to detect nuanced errors in reasoning steps. It consists of 6,216 problems with a total of 83,456 step-level labels, assessing models across multiple dimensions: Simplicity (non-redundancy, non-circular logic), Soundness (empirical soundness, step consistency, domain consistency, confidence invariance), and Sensitivity (prerequisite sensitivity, deception resistance, multi-solution consistency). The benchmark uses both synthetic and human-verified data, with rigorous quality control measures, including manual verification of a subset of data. A composite metric, PRMScore, is introduced, combining positive and negative F1 scores for a balanced evaluation.

To evaluate the effectiveness of \model{} in guiding step-by-step mathematical problem-solving, we employ it to assign scores to individual reasoning steps generated by large language models (LLMs), selecting only those with the highest overall rewards to build upon. This evaluation is conducted across several widely recognized math reasoning benchmarks, including AIME24, AMC23, MATH, Olympiad Bench, College MATH, and Minerva MATH\footnote{Following \cite{she2025r}, we use their subset of 200 test samples for Olympiad Bench and College MATH.}.

\subsection{Baselines}
Our evaluation utilizes a diverse set of discriminative process reward models from recent literature as baselines: \textit{Math-Shepherd}~\cite{Wang2024}, \textit{Math-PSA}~\cite{wang2024openr}, \textit{RLHFlow-Mistral and RLHFlow-DeepSeek}~\cite{xiong2024rlhflowmath}, \textit{Skywork-PRM-7b}~\cite{skywork2024skywork}, \textit{ReasonEval-7B}~\cite{xia2025evaluating}, \textit{Llemma-PRM800k-7B} ~\cite{sun2024easy}, \textit{Qwen2.5-Math-PRM-7B} and \textit{Qwen2.5-Math-7B-PRM800K}~\cite{zhang2025lessons}. We selected these baselines to cover a diverse range of training regimes, including models trained on human annotations, automated annotations, and hybrid approaches, as well as varying scales of training data. 

\begin{figure*}[ht!]
    \centering
    \includegraphics[width=\textwidth]{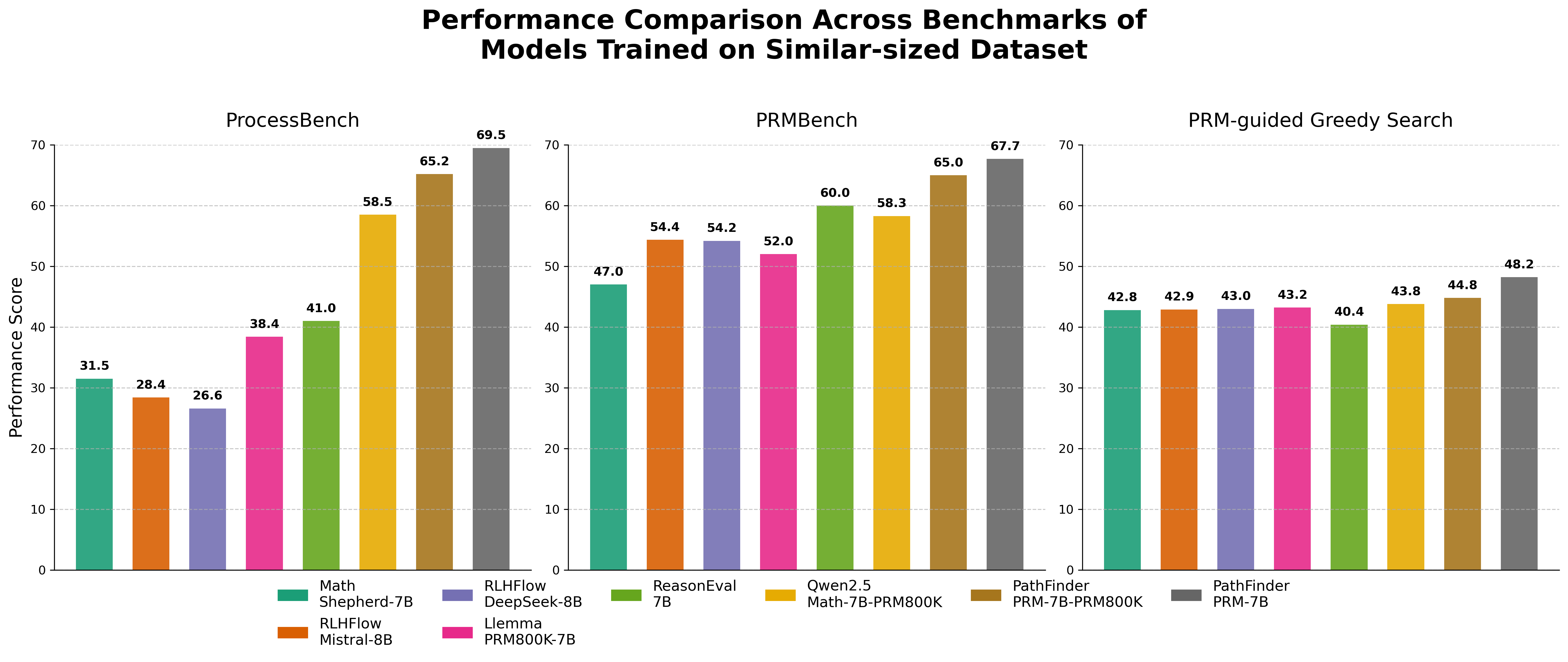}
    \caption{Performance comparison of language models across three benchmarks. The \model{}-7B model (gray) shows the highest performance across all benchmarks.}
    \label{fig:benchmarks}
\end{figure*}

\section{Results}

\begin{table*}[ht!]
\centering
\small
\begin{tabular}{lcccccc}
\toprule
\textbf{Model} & \textbf{\# Samples} & \textbf{GSM8K} & \textbf{MATH} & \textbf{Olympiad Bench} & \textbf{OmniMath} & \textbf{Avg. F1} \\
\midrule
\multicolumn{7}{l}{\textbf{Trained on Automated Annotation Data}} \\
Math-Shepherd-7B* & 445K & 47.9 & 29.5 & 24.8 & 23.8 & 31.5 \\
RLHFlow-Mistral-8B* & 273K & 50.4 & 33.4 & 13.8 & 15.8 & 28.4 \\
RLHFlow-DeepSeek-8B* & 253K & 38.8 & 33.8 & 16.9 & 16.9 & 26.6 \\
Qwen2.5-Math-PRM-7B* & $\sim$1.5M & \textbf{82.4} & \textbf{77.6} & \textbf{67.5} & \textbf{66.3} & \textbf{73.5} \\
\midrule
\multicolumn{7}{l}{\textbf{Trained on Human Annotated Data (PRM800K)}} \\
Llemma-PRM800K-7B & $\sim$350K & 48.4 & 43.1 & 28.5 & 33.4 & 38.4 \\
ReasonEval-7B$^\dag$ & $\sim$350K & 41.0 & 48.9 & 36.7 & 37.4 & 41.0 \\
Qwen2.5-Math-7B-PRM800K* & 264K & 68.2 & 62.6 & 50.7 & 44.3 & 58.5 \\
\model{-7B-PRM800K}  & $\sim$350K & \underline{74.1} & \underline{71.3} & 58.7 & 56.6 & \underline{65.2} \\
\hspace{0.5em} w/o Separate Subtask Prediction & & 71.4 & 71.1 & \underline{59.2} & \underline{58.0} & 64.9\\
\midrule
\multicolumn{7}{l}{\textbf{Trained on a Mix of Human and Automated Annotation Data}} \\
Math-PSA-7B$^\dag$  & $\sim$860K & 62.4 & 41.9 & 31.5 & 25.2 & 40.3 \\
Skywork-PRM-7B* & unk & 70.8 & 53.6 & 22.9 & 21.0 & 42.1 \\[1ex]
$\circledast$ \model{-7B} & $\sim$400K &  \underline{77.9} & \underline{75.3} & \underline{65.0} & \underline{59.7} & \underline{69.5} \\
\hspace{0.5em} w/o Separate Error Categories & & 76.1 & 73.8 & 61.4 & 56.6 & 67.0 \\
\hspace{0.5em} w/o Separate Subtask Prediction & & 73.9 & 72.6 & 63.9 & 59.9 & 67.6 \\
\bottomrule
\end{tabular}
\caption{Performance (F1) on ProcessBench. Results marked with * come from \citeauthor{zhang2025lessons}.The best performance across all categories is in \textbf{bold} and the best performance within a category is \underline{underlined}.\# Samples denotes the number of training samples used by each model.}
\label{tab:process_bench_main}
\end{table*}

\begin{table*}[ht!]
\centering
\resizebox{\textwidth}{!}{%
\begin{tabular}{@{}l *{7}{c}@{}}
\toprule
\textbf{Setting}
  & \textbf{AIME24} & \textbf{AMC23} & \textbf{MATH} & \textbf{Olympiad Bench} & \textbf{College MATH} & \textbf{Minerva MATH} & \textbf{Avg}. \\
\midrule
pass@1*               & 11.2 & 47.8 & 73.0 & 38.0 & 38.6 & 37.2 & 41.0 \\
major@8*              & 20.0 & 57.5 & 79.6 & 47.0 & 41.5 & 42.7 & 48.0 \\
pass@8* & 33.3 & 82.5 & 88.8 & 58.5 & 47.5 & 57.7 & 61.4 \\
\midrule
\multicolumn{8}{l}{{\textbf{Reward Guided Search (prm@8)}}} \\
\midrule
Math‑Shepherd‑7B*     & 13.3 & 52.5 & 74.6 & 38.5 & 36.5 & 41.2 & 42.8 \\
Math‑PSA‑7B*          &  6.7 & 57.5 & 79.8 & 42.5 & 41.0 & 39.3 & 44.5 \\
RLHFlow‑PRM‑Mistral‑8B*     & 10.0 & 57.5 & 73.4 & 37.5 & 38.0 & 41.2 & 42.9 \\
RLHFlow‑PRM‑DeepSeek‑8B*    & 13.3 & 52.5 & 74.8 & 39.5 & 37.0 & 40.8 & 43.0 \\
Lemma‑PRM800k‑7B*      & 13.3 & 57.5 & 73.8 & 40.0 & 36.5 & 38.2 & 43.2 \\
Skywork‑PRM‑7B*        & 10.0 & 57.5 & 77.8 & 41.5 & 39.0 & 43.4 & 44.9 \\
ReasonEval‑7B*         &  3.3 & 55.0 & 73.0 & 37.5 & 35.5 & 37.9 & 40.4 \\
Qwen2.5‑Math‑7B‑PRM800K* & \textbf{23.3} & 45.0 & 78.2 & 42.0 & 35.5 & 38.6 & 43.8 \\
Qwen2.5‑Math‑PRM‑7B*    & 16.7 & 60.0 & \textbf{81.0} & 43.5 & 39.0 & 40.4 & 46.8 \\
$\circledast$ \model{-7B-PRM800K} &  20.0 & 55.0 & 79.0 & 36.0 & 55.0 & 36.4 & 46.9 \\
\hspace{0.5em} w/o Separate Subtask Prediction &  6.6 & 55.0 & 82.2 & 36.0 & 53.5 & 36.0 & 45.0 \\
\midrule
$\circledast$ \model{-7B}&  20 & 62.5 & 78.8 & 36.5 & 55.0 & 36.7 & \textbf{48.3} \\
% \model{-7B}         &  ?? & ?? & ?? & ?? & ?? & ?? & ?? \\
\hspace{0.5em} w/o Separate Error Categories &  13.3 & 52.5 & 80.4 & 35.5 & 53.5 & 37.5 & 45.4 \\
\hspace{0.5em} w/o Separate Subtask Prediction &  10.0 & 55.0 & 81.6 & 37.0 & 53.5 & 36.0 & 45.5 \\
\bottomrule
\end{tabular}
}
\caption{The performance of PRM guided greedy search with Qwen2.5‑7B‑Instruct as the policy model. Results marked with * come from \citeauthor{she2025r}}
\label{tab:greedy-guide-main}
\end{table*}

\subsection{Main Results}

\paragraph{\model{-7B} is the SOTA for PRMBench:} \Cref{tab:prmbench_main} shows the PRMBench results of the selected baselines, \model{-variants} as well as LLM-as-judge performance of strong open-source and proprietary LLMs. In the discriminative PRM category, \model{-7B} achieves the highest overall PRM score (67.7), outperforming Qwen2.5-Math-PRM-7B (65.5) and ReasonEval-7B (60.0). The variant \model{-7B-PRM800K}, trained on a fraction of our dataset, achieves a competitive score of 65.0. Notably, \model{-7B} outperforms nearly all LLM-as-Judge models, including GPT-4o, QwQ-Preview-32B and Gemini-2.0-flash-exp. PRMBench is a benchmark designed to test a model's ability to detect subtle and complex errors. Our results affirm that our hierarchical PRM approach enables the model to detect these nuanced errors, leading to stronger process-level understanding and supervision.

\paragraph{\model{} Excels on ProcessBench:}
\Cref{tab:process_bench_main} presents F1 results on ProcessBench. When trained exclusively on PRM800K, \model{-7B-PRM800K} attains an average F1 of 65.2, beating the previous best (Qwen2.5‑Math‑7B‑PRM800K, 58.5) by 6.7 points and outperforming all other PRM800K‐only baselines across every category: GSM8K (+5.9), MATH (+8.7), Olympiad Bench (+8.0) and OmniMath (+12.3).  

Leveraging a larger, mixed human + auto‑annotated dataset further boosts performance. \model{-7B} achieves an average F1 of 69.5, setting new state–of–the–art among mixed‑data models and closing the gap to the top automated‑annotation model (Qwen2.5‑Math‑PRM‑7B*, 73.5) to just 4 points. Notably, \model{-7B} also leads in every individual benchmark—GSM8K (77.9), MATH (75.3), Olympiad Bench (65.0), and OmniMath (59.7), demonstrating the scalability and robustness of our hierarchical reward modeling approach.  

\paragraph{Improved Reward-Guided Search with Better PRMs:}
Finally, we assess the utility of our PRM in guiding solution search. Using Qwen2.5-Instruct-7B as a generator and ranking sampled steps in completions using our PRM, \Cref{tab:greedy-guide-main} shows that \model{-7B} yields the highest average prm@8 score (48.25), outperforming Qwen2.5-Math-PRM-7B (46.8). The advantage holds across tasks, including challenging subsets such as AIME24 and College MATH, indicating better inductive bias and alignment with ground-truth solution quality.

\paragraph{\model{} is Competitive to Qwen2.5‑Math‑PRM‑7B Despite Using \~3× Less Data:}
Although Qwen2.5‑Math‑PRM‑7B was trained on roughly 1.5M automated annotations, our \model{-7B}, trained on only $\sim$400K samples, matches or exceeds its performance in key benchmarks and reward-guided search. On ProcessBench, \model{-7B} performs competitively to Qwen2.5‑Math‑PRM‑7B in average F1 69.5 vs 73.5 despite using less than one-third of the data. More importantly, \model{-7B} surpasses Qwen2.5‑Math‑PRM‑7B on PRMBench overall (67.7 vs 65.5), and drives higher pass@8 in reward‑guided greedy search (48.3 vs. 46.8). This demonstrates that our hierarchical, error‑aware training yields more data‑efficient and robust PRMs, achieving superior process supervision with far fewer samples.

\begin{figure}[h!]
    \centering
    \includegraphics[width=0.9\columnwidth]{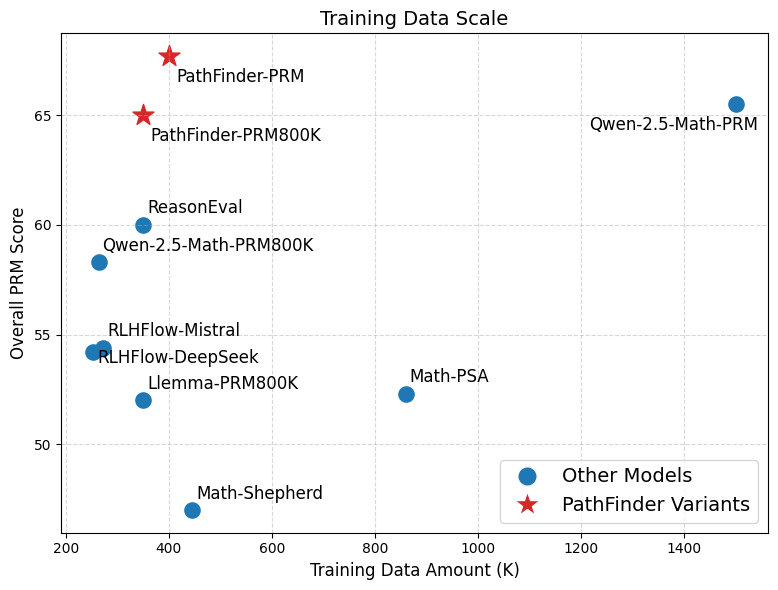}
    \caption{PRMBench Overall  PRMscore against the data scales of different baselines and \model{} variants.}
    \label{fig:data_scale}
\end{figure}

As shown in \Cref{fig:benchmarks}, when comparing against PRMs trained on similarly sized datasets, \model{} consistently achieves superior performance across all benchmarks. \Cref{fig:data_scale} presents the performance of various PRMs on PRMBench. The results demonstrate that \model{} not only surpasses other PRMs trained on comparably sized datasets but also outperforms Math-PSA and Qwen-2.5-Math-PRM, despite those models being trained on 2–3 times more data.

\subsection{Ablations}
In our approach, we made two main claims: (1) decoupling the subtasks of Error Detection and Correct Path Guidance, and (2) categorizing errors into two prominent error categories in math will boost PRM performance. To verify these claims, we performed ablation experiments by modifying parts of our method:
\begin{enumerate}
    \item \model{} \textbf{w/o Separate Subtask Prediction}: Following existing PRM approaches, we trained the model to jointly learn to tackle both error detection and correct path guidance using only the step correctness labels.
    
    \item \model{} \textbf{w/o Separate Error Categories}: In this approach, we still do a hierarchical prediction but we modify step 1. Instead of detecting the presence of 2 error categories, we combined the categories and predicted the presence of an error in the step.
\end{enumerate}

\paragraph{\model{} Benefits from Separating Error Categories:}
On ProcessBench, explicitly distinguishing math and consistency errors yields a clear overall boost: \model{-7B} scores 69.5 Avg. F1, versus 67.0 for the \model{-7B} w/o Separate Error Categories. We also observe a similar drop in performance on PRMBench, the \model{ w/o Separate Error Categories}  shows a small drop in performance (0.4 points) compared to \model{-7B}

Crucially, reward‑guided search highlights the practical impact of error typing: when ranking eight candidate solutions, \model{-7B} achieves 48.3 prm@8, compared to just 45.4 for \model{ w/o Separate Error Categories} (+2.9 points). This jump in real‑world problem‑solving performance highlights that fine‑grained error signals not only improve diagnostic metrics but can also translate directly into selecting higher‑quality solution paths.

\paragraph{\model{} Benefits from Error-Aware Hierarchical Supervision:}
Across ProcessBench, PRMBench, and reward-guided search, \model{} consistently outperforms the \model{} w/o separate subtask prediction, demonstrating the importance of hierarchical modeling of the subtasks. On ProcessBench, \model{-7B} improves from 67.6 to 69.5 F1 (+1.9), and on PRMBench, from 64.9 to 67.7 (+2.8). In reward-guided search, the improvement is similarly clear: 48.3 prm@8 versus 45.5. These results highlight the value of decoupling feedback prediction into discrete reasoning components.

\paragraph{Scalable Performance with Additional Training Data:} Training on an additional 50K samples from a broader, automatically annotated dataset greatly boosted the performance of \model{} and helped it reach state-of-the-art performance. \model{-7B} outperforms \model{-7B-PRM800K} across ProcessBench, PRMBench, and reward-guided greedy search, demonstrating the benefits of scaling beyond PRM800K dataset. On ProcessBench, \model{-7B} achieves 69.5 Avg.F1 versus 65.2, while on PRMBench it improves from 65.0 to 67.7 F1. In reward-guided search, \model{-7B} raises pass@8 from 46.9 to 48.3. These gains highlight that error-aware hierarchical modeling scales well with increased data diversity and quantity, enabling stronger generalization and robustness. 
\fbox{\parbox{\columnwidth}{A key distinction is that the 50K additional data was collected using Mistral-7B as a generator, while PRM800K was produced using a fine-tuned GPT-4 model. Exposure to reasoning traces from a smaller and weaker model may have helped \model{} better learn to recognize and correct common failure patterns, contributing to improved robustness and generalization across benchmarks. We leave a deeper investigation of this hypothesis to future work.}}

\section*{Conclusion}
In this work, we introduced \model{}, a hierarchical discriminative process reward model that decouples error detection from step optimality guidance, classifying math and consistency errors before computing step rewards. Our evaluation and ablation trials demonstrate that error-aware hierarchical supervision yields notable improvements on PRM benchmarks, with \model{-7B} achieving state-of-the-art performance among discriminative PRMs on PRMBench and strong results on ProcessBench despite using less than three times the data used to train the current best performing model. A similar performance boost is observed in reward-guided search evaluation, affirming our hypothesis about the efficacy of hierarchical error-aware reward generation. Our approach is, therefore, a promising direction for more robust and interpretable process reward models, with potential for further gains when scaled to larger architectures.

\section*{Limitations}
Due to computational constraints, our experiments were limited to 7B models. While this scale provides a strong foundation for evaluating our proposed methodology, we hypothesise that larger models could further enhance modeling accuracy and better leverage process supervision signals due to their improved mathematical reasoning capabilities.

\bibliography{custom}

\newpage
\onecolumn
\appendix

\section{Extended Results}
\label{sec:extended_results}

\subsection{ProcessBench}
\Cref{tab:process_bench_full} contains the extended evaluation on ProcessBench with LLM-as-Judge Baselines using both Proprietary and Open-Source Language Models. \\ 

\begin{table*}[ht]
\centering
\small
\begin{tabular}{lcccccc}
\toprule
\textbf{Model} & \textbf{\# Samples} & \textbf{GSM8K} & \textbf{MATH} & \textbf{Olympiad Bench} & \textbf{OmniMath} & \textbf{Avg. F1} \\
\midrule
\multicolumn{7}{l}{\textbf{\textit{LLM-as-judge, Proprietary language models}}} \\
\midrule
GPT-4o* & unk & 79.2 & 63.6 & 51.4 & 53.5 & 61.9 \\
o1-mini* &  unk & \textbf{93.2} & 88.9 & 87.2 & 82.4 & 87.9 \\
\midrule
\multicolumn{7}{l}{\textbf{\textit{LLM-as-judge, Open-source language models}}} \\
\midrule
Llama-3.3-70B-Instruct* & unk & 82.9 & 59.4 & 46.7 & 43.0 & 58.0 \\
Qwen2.5-Math-72B-Instruct* &  unk & 65.8 & 52.1 & 32.5 & 31.7 & 45.5 \\
Qwen2.5-72B-Instruct* &  unk & 76.2 & 61.8 & 54.6 & 52.2 & 61.2 \\
\midrule
\multicolumn{7}{l}{\textbf{\textit{Discriminative Process Reward Models}}} \\
\midrule
\multicolumn{7}{l}{\textbf{Trained on Automated Annotation Data}} \\
Math-Shepherd-7B* & 445K & 47.9 & 29.5 & 24.8 & 23.8 & 31.5 \\
RLHFlow-Mistral-8B* & 273K & 50.4 & 33.4 & 13.8 & 15.8 & 28.4 \\
RLHFlow-DeepSeek-8B* & 253K & 38.8 & 33.8 & 16.9 & 16.9 & 26.6 \\
Qwen2.5-Math-PRM-7B* & $\sim$1.5M & \textbf{82.4} & \textbf{77.6} & \textbf{67.5} & \textbf{66.3} & \textbf{73.5} \\[1ex]
\multicolumn{7}{l}{\textbf{Trained on Human Annotated Data (PRM800K)}} \\
Llemma-PRM800K-7B & $\sim$350K & 48.4 & 43.1 & 28.5 & 33.4 & 38.4 \\
ReasonEval-7B$^\dag$ & $\sim$350K & 41.0 & 48.9 & 36.7 & 37.4 & 41.0 \\
Qwen2.5-Math-7B-PRM800K* & 264K & 68.2 & 62.6 & 50.7 & 44.3 & 58.5 \\
\model{-7B-PRM800K}  & $\sim$350K & \underline{74.1} & \underline{71.3} & 58.7 & 56.6 & \underline{65.2} \\
\hspace{0.5em} w/o Separate Subtask Prediction & & 71.4 & 71.1 & \underline{59.2} & \underline{58.0} & 64.9\\[1ex]
\multicolumn{7}{l}{\textbf{Trained on a Mix of Human and Automated Annotation Data}} \\
Math-PSA-7B$^\dag$  & $\sim$860K & 62.4 & 41.9 & 31.5 & 25.2 & 40.3 \\
Skywork-PRM-7B* & unk & 70.8 & 53.6 & 22.9 & 21.0 & 42.1 \\[1ex]
$\circledast$ \model{-7B} & $\sim$400K &  \underline{77.9} & \underline{75.3} & \underline{65.0} & \underline{59.7} & \underline{69.5} \\
\hspace{0.5em} w/o Separate Error Categories & & 76.1 & 73.8 & 61.4 & 56.6 & 67.0 \\
\hspace{0.5em} w/o Separate Subtask Prediction & & 73.9 & 72.6 & 63.9 & 59.9 & 67.6 \\
\bottomrule
\end{tabular}
\caption{Performance (F1) on ProcessBench. Results marked with * come from \citeauthor{zhang2025lessons}. The best performance across all categories is in \textbf{bold} and the best performance within a category is \underline{underlined}.}
\label{tab:process_bench_full}
\end{table*} 

\section{Scaling Effects of Additional RLHFlow Mistral Data}

To further probe the effects of training data scaling, we increased the amount of RLHFlow Mistral data from 50K to 200K samples and retrained \model{}. As shown in \Cref{tab:data_scale_results}, this increase from 400K to 550K total training samples did not result in improved performance. In fact, we observe a slight drop across all benchmarks: ProcessBench performance decreased from 69.5 to 68.75 Avg.F1, PRMBench from 67.7 to 67.4, and reward-guided search accuracy dropped from 48.3 to 47.5.

These findings suggest that the benefits of augmenting training with weaker model-generated traces (e.g., from Mistral-7B) may saturate quickly. Simply increasing the volume of such data does not necessarily lead to improved generalization, and may even slightly degrade performance. This underscores the importance of data quality and the nuanced role of diversity over quantity in training PRMs. We leave further study into effective feedback curation and dataset composition to future work.

\begin{table*}[ht!]
\centering
\small
\resizebox{\textwidth}{!}{%
\begin{tabular}{lcccccc}
\toprule
\textbf{Model} & \textbf{\# Total samples} & \textbf{\# Mistral samples} & \textbf{ProcessBench} & \textbf{PRMBench} & \textbf{Reward Guided Search (PRM@8)} \\
\midrule
\model{-7B-PRM800K} & 350K & 0 & 65.2 & 65.0 & 46.9 \\
\model{-7B} & 400K & 50K & 69.5 & 67.7 & 48.3 \\
\model{-7B} & 550K & 200K & 68.75 & 67.4 & 47.5 \\
\bottomrule
\end{tabular}
}
\caption{Effect of scaling RLHFlow Mistral data on \model{} performance.}
\label{tab:data_scale_results}
\end{table*}

\section{Design Choice: Two Forward Passes over Autoregressive Multi-Token Prediction}
\label{sec: inference design}

In \model{}, we adopt a two-forward-pass approach to predict intermediate error labels—Math Error and Consistency Error—followed by a final reward score for correctness. This contrasts with autoregressive decoding, which would generate all three labels sequentially. The motivation behind this design is to minimize error cascading between dependent predictions. Specifically, we aim to predict Math and Consistency labels independently in the first forward pass to prevent the Consistency prediction from being influenced by the previously generated Math label. This is a potential issue in the autoregressive setup since each token depends on the ones before it.

In our setup, a single language modeling head is used without any task-specific heads. During the first forward pass, we construct the input as:
\texttt{[PRM Input], Math: <mask>, Consistency: <mask>},
where the two special \texttt{<mask>} tokens represent the target positions for Math and Consistency labels. The model is trained to produce the correct label tokens at these masked positions, allowing us to decode both predictions simultaneously without one influencing the other.

In the second forward pass, we supply the previously predicted Math and Consistency tokens in place, and append a third \texttt{<mask>} token to infer the final correctness label:
\texttt{[PRM Input], Math: predicted\_label, Consistency: predicted\_label, Correctness: <mask>}.
The probability corresponding to the positive reward token at the final mask position is then used as the model's output reward score.

This inference method avoids sequential decoding and uses only masked forward passes, enabling clearer modular supervision and avoiding implicit dependency leakage between intermediate labels. During training, a similar masked format is used. This clean separation helps \model{} to better capture independent error signals and contributes to its robustness.

\begin{algorithm}[H]
\caption{\model{} Inference via Two Forward Passes}
\label{alg:two_pass_inference}
\begin{algorithmic}[1]
\REQUIRE Prompt P, \model{}
\vspace{0.5em}
\STATE // First Forward Pass: Predict Math and Consistency Labels
\vspace{0.2em}

\STATE $Input_1 \gets concatenate(P, \text{"Math: <mask>"}, \text{"Consistency: <mask>"})$
\STATE $Logits_1 \gets \model{}.forward(Input_1)$
\STATE $pred_{math} \gets argmax(Logits_1 \text{ at Math mask position})$
\STATE $pred_{consistency} ← argmax(Logits_1 \text{ at Consistency mask position}$)

\vspace{1em}

\STATE // Second Forward Pass: Predict Correctness/Reward Label

\vspace{0.4em}

\STATE \parbox[t]{\dimexpr\linewidth-\algorithmicindent}{
  $Input_2 \gets concatenate(P, \text{"Math: "}, pred_{math},$\\
  $\phantom{Input_2 \gets concatenate} \text{"Consistency: "}, pred_{consistency},$\\
  $\phantom{Input_2 \gets concatenate} \text{"Correctness: <mask>"})$
}
\vspace{1em}
\STATE $Logits_2 \gets \model{}.forward(Input_2)$
\STATE $reward_{prob} \gets softmax(Logits_2 \text{ at Correctness mask position})$
\STATE $pred_{reward} \gets reward_{prob} \text{ of <+> token}$  
\vspace{0.4em}
\RETURN $pred_{math}, pred_{consistency}, pred_{reward}$
\end{algorithmic}
\end{algorithm}

\newpage{}

\section{Data Annotation Prompt}
The prompt below was utilized with DeepSeek-R1-Distill-Qwen-32B to synthesize the 3-dimensional categorical score vectors assigned to each sample in the dataset used to train \model{} 

\begin{promptbox}[Prompt for dataset labelling]
You are an analytical math instructor grading a student's work. Think step-by-step through your analysis. Below is the math question, the previous steps by the student, and the current step to evaluate. \\

\{context\} \\

Your task is to rigorously examine the current step and determine if it contains ANY mathematical errors. Assign binary scores (0 = wrong, 1 = correct) based on three criteria: \\

A) Mathematical logic — Is the current step, **on its own**, mathematically valid? Check for:
   • Calculation errors
   • Incorrect formula application 
   • Invalid operations or simplifications
   • Algebra mistakes or sign errors
   • Incorrect assertions \\

B) Consistency — Is the current step logically consistent with:
   • Established ground truth
   • Previous steps
   • Any constraints or conditions established earlier
   • The mathematical domain applicable to this problem\\

C) Simplicity and optimality — is this step an efficient next step toward the solution? Check for:
   • Redundant statements: factually correct statements that do not help progress toward the solution.
   • Circular logic: does this step come to a conclusion already previously established?
   • Non-clarity: Are the assertions made in this step ambiguous in a way that obsfucates their purpose?
   • Optimality: is the **idea** of this step the near optimal approach one would take to solve the problem? \\

Double check all listed criterion here explicitly in your reasoning. In your analysis, be sensitive to subtle issues like missing pre-requisites/assumptions, correct-looking statements with slight errors and high confidence statements containing errors. \\

IMPORTANT POINTS: 

- If you find ANY error, even a minor one, you MUST assign a score of 0 to the appropriate criteria. Be skeptical and verify all claims thoroughly.

- For incorrect steps, wherever possible, attempt to categorize the issue as violating **one of the three criterion** (i.e., assign score 0 to **only one category**). Assign multiple 0 scores only for serious errors. \\

You must format your answer as below: \\

Reasoning: \\
\{\{Provide detailed analysis, showing all verification steps and explicitly identifying any errors found\}\} \\

Final answers: \\
Score A: \\
\{\{0 or 1 only\}\} \\
Score B: \\
\{\{0 or 1 only\}\} \\
Score C: \\
\{\{0 or 1 only\}\}

\end{promptbox}

\section{Training HyperParameters}
\Cref{tab:training-config} contains the hyperparameters used to train \model{}. We used the Transformers library Trainer implementation to train our model in a seq-to-seq manner \cite{wolf-etal-2020-transformers}.

\begin{table}[h]
\centering
\begin{tabular}{ll}
\toprule
\textbf{Parameter} & \textbf{Value} \\
\midrule
Model & Qwen\/Qwen2.5-Math-7B-Instruct \\
Torch Data type & bfloat16 \\
Attention Implementation & flash attention 2 \\
Per-device Train Batch Size & 2 \\
Gradient Accumulation Steps & 32 \\
Learning Rate & 1.0e-05 \\
Number of Training Epochs & 4 \\
LR Scheduler Type & cosine \\
Max Gradient Norm & 1.0 \\
Warmup Ratio & 0.1 \\
Seed & 42 \\
BF16 & true \\
Optimizer & adam \\
Gradient Checkpointing & True \\
\bottomrule
\end{tabular}
\caption{Training Configuration for Qwen2.5-Math-7B-Instruct.}
\label{tab:training-config}
\end{table}

\end{document}